\documentclass[10.5pt,compsoc]{BD}
\usepackage{graphicx}
\usepackage{footmisc}
\usepackage{subfigure}
\usepackage{url}
\usepackage{multirow}
\usepackage[noadjust]{cite}
\usepackage{amsmath,amsthm}
\usepackage{amssymb,amsfonts}
\usepackage{booktabs}
\usepackage{color}
\usepackage{ccaption}
\usepackage{booktabs}
\usepackage{float}
\usepackage{fancyhdr}
\usepackage{caption}
\usepackage{xcolor,stfloats}
\usepackage{comment}
\graphicspath{{figures/}}
\usepackage{cuted}  
\usepackage{captionhack}
\usepackage{epstopdf}

\newcommand{\upcite}[1]{\textsuperscript{\textsuperscript{\small\cite{#1}}}}

\usepackage[numbers,sort&compress]{natbib}
\usepackage[T1]{fontenc}

\usepackage[utf8]{inputenc}

\usepackage{microtype}

\usepackage{inconsolata}

\usepackage[commandnameprefix=always]{changes}

\usepackage{amsfonts}
\usepackage[scaled]{helvet} 
\usepackage{courier}
\usepackage{url}
\usepackage{float,color}
\usepackage{times}
\usepackage{algorithm}
\usepackage{algorithmic}
\usepackage{float}
\usepackage{pdfpages}
\usepackage{changes}
\usepackage{epsfig}
\usepackage{stfloats}
\usepackage{url}
\usepackage{diagbox}
\usepackage{subfigure}
\usepackage{epstopdf}
\usepackage{multicol}
\usepackage{makecell}
\usepackage{longtable}
\usepackage{dsfont,amsfonts,color}
\usepackage{multirow}
\usepackage{booktabs}
\usepackage{svg}
\usepackage{xr}
\usepackage{tablefootnote}

\csname url@samestyle\endcsname

\providecommand{\BIBentrySTDinterwordspacing}{\spaceskip=0pt\relax}
\providecommand{\BIBentryALTinterwordstretchfactor}{4}
\providecommand{\BIBentryALTinterwordspacing}{\spaceskip=\fontdimen2\font plus
\BIBentryALTinterwordstretchfactor\fontdimen3\font minus \fontdimen4\font\relax}
\providecommand{\BIBforeignlanguage}[2]{{%
\expandafter\ifx\csname l@#1\endcsname\relax
\typeout{** WARNING: IEEEtran.bst: No hyphenation pattern has been}%
\typeout{** loaded for the language `#1'. Using the pattern for}%
\typeout{** the default language instead.}%
\else
\language=\csname l@#1\endcsname
\fi
#2}}
\providecommand{\BIBdecl}{\relax}
\BIBdecl
\headevenname{\zihao{-5}{\textbf{\emph{Big Data Mining and Analytics, April}}} 2024, 1(1): 000-000}%
\headoddname{\zihao{-5}{\sf Ziqian Zeng et al.:}\quad {\textbf{\emph{Chimera: A Lossless Decoding Method for Accelerating Large Language
Models Inference by Fusing all Tokens}}}}%

\newtheoremstyle{mystyle}{0pt}{0pt}{\normalfont}{1em}{\bf}{}{1em}{}
\theoremstyle{mystyle}
\renewcommand\figurename{Fig.~}

\newcommand{\nop}[1]{}

\def\t{{\bf t}}

\def\0{{\bf 0}}
\def\1{{\bf 1}}

\def\RM{{\mathcal R}}

\def\MM{{\mathcal M}}

\def\RM{{\mathcal R}}

\def\TM{{\mathcal T}}

\makeatletter
\renewcommand{\@biblabel}[1]{[#1]\hfill}
\makeatother

\begin{document}




\hyphenpenalty=50000

\makeatletter
\newcommand\mysmall{\@setfontsize\mysmall{7}{9.5}}

\newenvironment{tablehere}
  {\def\@captype{table}}
  {}
\newenvironment{figurehere}
  {\def\@captype{figure}}
  {}

\thispagestyle{plain}%
\thispagestyle{empty}%

\let\temp\footnote
\renewcommand \footnote[1]{\temp{\zihao{-5}#1}}
{}
\vspace*{-40pt}
\noindent{\zihao{5-}\textbf{\scalebox{0.95}[1.0]{\makebox[5.9cm][s]
{BIG\hfill  DATA \hfill MINING \hfill AND \hfill ANALYTICS}}}}

\vskip .2mm
{\zihao{5-}
\textbf{
\hspace{-5mm}
\scalebox{1}[1.0]{\makebox[5.6cm][s]{%
I\hspace{0.70pt}S\hspace{0.70pt}S\hspace{0.70pt}N\hspace{0.70pt}{\color{white}%
2\hspace{-2pt}2\hspace{0.70pt}}2\hspace{0.70pt}0\hspace{0.70pt}9\hspace{0.70pt}6\hspace{0.70pt}-\hspace{0.70pt%
}0\hspace{0.70pt}6\hspace{0.70pt}5\hspace{0.00pt}4\hspace{0.70pt}\hspace{0.70pt%
}\hspace{0.70pt}{\color{white}l\hspace{0.70pt}l\hspace{0.70pt}}0\hspace{0.70pt}%
?\hspace{0.70pt}/\hspace{0.70pt}?\hspace{0.70pt}?\hspace{0.70pt}{\color{white}%
l\hspace{0.70pt}l\hspace{0.70pt}}p\hspace{0.70pt}p\hspace{0.70pt}?\hspace{0.70pt}?\hspace{0.70pt}?%
--\hspace{ 0.70pt}?\hspace{0.70pt}?\hspace{0.70pt}?}}}

\vskip .2mm\noindent
{\zihao{5-}\textbf{\scalebox{1}[1.0]{\makebox[5.6cm][s]{%
V\hspace{0.4pt}o\hspace{0.4pt}l\hspace{0.4pt}u\hspace{0.4pt}m\hspace{0.4pt}%
e\hspace{0.4em}1\hspace{0.4pt},\hspace{0.8em}N\hspace{0.4pt}u\hspace{0.4pt}%
m\hspace{0.4pt}b\hspace{0.4pt}e\hspace{0.4pt}r\hspace{0.4em}1,\hspace{0.8em}%
J\hspace{0.4pt}a\hspace{0.4pt}n\hspace{0.4pt}u\hspace{0.4pt}a\hspace{0.4pt}%
\hspace{0.4pt}r\hspace{0.4pt}y\hspace{0.4em}2\hspace{0.4pt}0\hspace{0.4pt}1\hspace{0.4pt}8}}}}


\vskip .2mm\noindent
{\zihao{5-}\textbf{\scalebox{1}[1.0]{\makebox[5.6cm][s]{%
\color{white}{V\hfill o\hfill l\hfill u\hfill m\hfill%
e\hspace{0.356em}1,\hspace{0.356em}N\hfill u\hfill%
m\hfill b\hfill e\hfill r\hspace{0.356em}1,\hspace{0.356em}%
S\hfill e\hfill p\hfill t\hfill e\hfill%
m\hfill b\hfill e\hfil lr\hspace{0.356em}2\hfill0\hfill1\hfill8}}}}}\\

\begin{strip}
{\center
{\zihao{3}\textbf{Chimera: A Lossless Decoding Method for Accelerating Large Language Models Inference by Fusing all Tokens}}
\vskip 9mm}

{\center {\sf \zihao{5}
 Ziqian Zeng$^{\dag}$, Jiahong Yu$^{\dag}$, Qianshi Pang, Zihao Wang, Huiping Zhuang, Hongen Shao, Xiaofeng Zou*

}
\vskip 5mm}
%

\centering{
\begin{tabular}{p{160mm}}

{\zihao{-5}
\linespread{1.6667} %
\noindent
{\fontfamily{phv}\selectfont
\textbf {Abstract}:} {\sf
Large language models (LLMs) have demonstrated remarkable capabilities across various tasks. However, their widespread application is hindered by the resource-intensive decoding process. To address this challenge, current approaches have incorporated additional decoding heads to enable parallel prediction of multiple subsequent tokens, thereby achieving inference acceleration. Nevertheless, the accuracy of these decoding heads falls short of the auto-regressive decoding approach.In light of these limitations, we propose Chimera, a novel framework specifically designed for speculative sampling. Within this framework, we introduce a lightweight draft model that effectively utilizes previously generated tokens to predict subsequent words. To ensure both accuracy and efficiency, we present two strategies within the lightweight draft model. Firstly, we focus on capturing short-range dependencies at the bottom layer. Secondly, we leverage the readily available representations from the original LLM.Through empirical evaluation on the Vicuna and LlaMA-2-chat series, Chimera demonstrates impressive results, achieving an average latency speedup ratio of 2.7x compared to the vanilla auto-regressive decoding approach. This highlights the potential of our proposed framework in significantly improving the efficiency of large language models during the decoding process.}
\vskip 4mm
\noindent
{\fontfamily{phv}\selectfont
\textbf{Key words}:} {\sf lossless decoding ; inference accelerating ; large language models}}

\end{tabular}
}
\vskip 6mm

\vskip -3mm
\zihao{6}\end{strip}

\thispagestyle{plain}%
\thispagestyle{empty}%
\makeatother
\pagestyle{tstheadings}

\begin{figure}[b]
\vskip -6mm
\begin{tabular}{p{44mm}}
\toprule\\
\end{tabular}
\vskip -4.5mm
\noindent
\setlength{\tabcolsep}{1pt}
\begin{tabular}{p{1mm}p{79.5mm}}
$\bullet$& Ziqian Zeng, Jiahong Yu, Qianshi Pang, Huiping Zhuang, Hongen Shao, Xiaofeng Zou are with the  South China University of Technology, Guangzhou, 510641, China. 
E-mail: zqzeng@scut.edu.cn; jiahongyugz@gmail.com; qianshipang@gmail.com; hpzhuang@scut.edu.cn; ftshaohongen@mail.scut.edu.cn; zouxiaofeng@scut.edu.cn \\
$\bullet$&  Zihao Wang is with The Hong Kong University of Science and Technology,  E-mail:zwanggc@cse.ust.hk \\

$\sf{*}$&
Correspondence Author \\
 $\dag$         &   Ziqian Zeng and Jiahong Yu     contribute equally to this work.\\
& Manuscript received: 2024-04-20 \\

\end{tabular}
\end{figure}\zihao{5}

\vspace{3.5mm}
\vspace{3.5mm}
\section{Introduction}
\label{s:introduction}
\noindent
The convergence of the cyber, physical, and social worlds has given rise to the concept of Cyber Physical Social Intelligence (CPSI), which enables the development of smart and interconnected systems across various domains. 
A key component of CPSI is the massive and complex Big Data that emerges from these integrated environments. 
Harnessing the value of this CPSI Big Data requires the employment of advanced Big Data computing techniques. 

One critical challenge in CPSI Big Data computing is the need for efficient and high-performance inference of large language models (LLMs)\upcite{almazrouei2023falcon, touvron2023llama2, achiam2023gpt4, anil2023palm, bai2023qwen, yang2023baichuan} , which play a pivotal role in extracting meaningful insights and powering intelligent decision-making within CPSI systems. 
LLMs, with their extensive knowledge and complex architectures, often face the high inference latency  during inference, posing a bottleneck in the real-time or near real-time response required by CPSI applications.

This latency arises partially from auto-regressive decoding, wherein the token produced at each step depends on all previously generated tokens. 
To alleviate the latency issue inherent to  auto-regressive decoding, a novel decoding approach called speculative decoding\upcite{stern2018blockwise, Leviathan2023fast} has emerged.

The process of speculative decoding\upcite{stern2018blockwise, 
Leviathan2023fast} can be summarized as "Draft-then-Verify." It involves two models: the original LLM and the draft model. 
The draft model first efficiently generates multiple speculations which consists of multiple subsequent tokens.
Subsequently, the original LLM verify all of these speculations to accept the one that meets the LLM’s verification criterion.
The process of drafting and verification is iterated until the termination condition is met. 
Given that the draft model\upcite{Leviathan2023fast,chen2023accelerating} is typically smaller in size than the original LLM, it can generate multiple tokens fast, thus achieving substantial acceleration.

However, the acquisition and maintenance of a separate draft model is challenging in real-world implementations. 
To address this issue, Blockwise Parallel Decoding\upcite{stern2018blockwise} and Medusa\upcite{cai2024medusa} added extra decoding heads on top of the last Transformer block to predict multiple subsequent tokens in parallel. 
Although these methods are simple and efficient, the accuracy of the decoding heads is significantly lower than the auto-regressive decoding head, which is supported by the findings presented in Table \ref{tab:head_acc}. 
This disparity in accuracy arises from the fact that LLMs are trained to predict the next word based on all previously generated words, whereas the decoding heads predict the next word based on the prefixed sequence lacking access to the tokens generated by preceding decoding heads. 
Specifically, suppose the draft model is going to generate speculated sequence $\{i+1, \cdots, i+k\}$ given the prefixed sequence $\{1, \cdots, i\}$, the parallel decoding head predicts the $(i+k)$-th word based on the prefixed sequence, without having access to the tokens $\{i+1, \cdots, i+k-1\}$.
The limited access to the entire context contributes to the lower accuracy observed in parallel decoding.


The acceleration effect of speculative decoding primarily depends on the number of drafted tokens accepted per verification. 
Improving the accuracy and speed of the draft model is one of the key factors in enhancing the acceptance rate\upcite{xia2024unlocking}. 
To enhance the accuracy of the draft model without significantly compromising speed, we propose a lightweight draft model that efficiently leverages all previously generated tokens to predict the next word. 

It is challenging to develop a lightweight draft model with a capacity comparable to the original LLM. 
We propose two strategies to cope with the challenge.

\textbf{The first strategy} is to capture short-range dependencies instead of long-range ones at the bottom layer of the lightweight draft model. 
As shown in previous work\upcite{jawahar2019what,rogers2020primer}, lower layers of pre-trained language models mainly focus on short-range dependencies, while deeper layers are capable of capturing long-range dependencies. 
Only capturing short-range dependencies can reduce inference time which is highly related to the number of tokens in the input sequence. 
Hence, we propose a trigram encoder to encode short-range dependencies, i.e., trigrams in the input sequence. 
An additional advantage of the trigram encoder is that we can retrieve the hidden states of trigrams through a pre-computed lookup table, instead of computing them on the fly, further saving inference time.

\textbf{The second strategy} is taking advantage of the readily available representation of the prefix sequence of the original LLM.
Capturing long-range dependencies across the entire context typically requires multiple Transformer blocks, but a lightweight draft model cannot afford that. 
Fortunately, the original LLM has properly encoded dependencies among prefix sequences. 
Therefore, we fully utilize the readily available hidden states from the original LLM. 
We propose a full context encoder implemented via a Transformer block to capture the long-range dependencies. 
The input of the full context encoder is the concatenation of mature hidden states from the original LLM and green hidden states from the trigram encoder. 
Finally, we use multiple residual decoding heads to generate subsequent words in different offset positions. 
Due to the limited capacities of the lightweight draft model, residual decoding heads not only take the hidden state of the last token from the full context encoder as input but also consider the hidden state of the last token from the original LLM.

The contributions of our method are summarized as follows,

$\bullet$ To improve the accuracy of the draft model, we propose a lightweight draft model that can consider all previous tokens when generating the next word. 

$\bullet$ We propose two strategies within the lightweight draft model to ensure accuracy and efficiency, i.e., capturing short-range dependencies at the bottom layer and taking advantage of readily available representation from the original LLM. 

$\bullet$ The experimental results show that our method can accelerate the vicuna-7b and vicuna-13b by 2.7x. 

The Chimera code is available at: \url{https://github.com/kafkayu/Chimera}


\section{Related Work}
\label{s:Related Work}


\noindent Drafting is crucial in the ``Draft-then-Verify'' pipeline of speculative decoding. 
By how the drafting process is conducted, the drafting methods can be roughly categorized into independent drafting and self-drafting.

\textbf{Independent Drafting.}
SpecDec\upcite{xia2023speculative} proposed employing an independent model for the drafting process. 
Furthermore, Recent research has emphasized diverse knowledge distillation methods to refine small LMs into proficient drafters, enhancing behavior alignment~\upcite{miao2024specinfer, kim2023speculative, zhou2023distillspec}. 
However, this method requires a wide range of computing resources to tune the drafting model. 
To alleviate the computation cost, a more direct and effective method is to use the same series of small LMs as drafters to speed up inference of their larger counterparts\upcite{Leviathan2023fast, spector2023accelerating, sun2024spectr, chen2023cascade}. 
However, a significant behavioral gap still exists between the small LM and the target LLM, resulting in suboptimal inference accuracy.

\textbf{Self-Drafting.}
To tackle the aforementioned challenges, certain approaches have been proposed to utilize the target LLM itself for efficient drafting\upcite{stern2018blockwise,santilli-etal-2023-accelerating, hooper2024speed}. 
Specifically, methods like Blockwise Decoding\upcite{stern2018blockwise} and Medusa\upcite{cai2024medusa} have introduced additional FFN heads on the top of the transformer decoder, allowing for the generation of multiple tokens simultaneously per step. Another line of research has leveraged early existing or layer-skipping techniques within the target LLM itself to handle the drafting task\upcite{yang2023predictive,zhang2023draft,monea2023pass}. For instance\upcite{yang2023predictive} , introduced additional subprocesses early in the current decoding step to initiate drafting future tokens in advance. 


\section{Background}
\subsection{Auto-regressive Decoding and Speculative Decoding}
\label{sec:autoregressive}


\noindent Autoregressive sampling is a common method for sequentially generating sequence data. 
In autoregressive sampling, the model generates sequence elements from left to right, one position at a time. 
After generating an element, it serves as the context condition for generating the next element, and this process is recursively repeated until the complete sequence is generated. 
The algorithm for autoregressive sampling can be described using the following formula:
\begin{equation}
    x_{t+1}\sim p_{t+1}=\mathcal{M}_{p}\left(x\mid x_{<t+1}\right),
\end{equation}
where $x_{t+1}$ is the predicted ($t+1$)-th word, $p_{t+1}$ is the predicted probability distribution of ($t+1$)-th word, $\MM(\cdot)$ is the LLM. 

Speculative decoding is a decoding paradigm where, at each decoding step, multiple future tokens are efficiently drafted first and then verified in parallel using the original LLM. 
Specifically, in speculative decoding, the process is divided into two steps: drafting and verification.

\textbf{Drafting.} During the drafting step, multiple future tokens are quickly generated in parallel without waiting for verification. 
This drafting process is typically more efficient as it can leverage parallelism and exploit the computational power of modern hardware.

\textbf{Verification.} Once the drafting step is completed, all the drafted tokens are verified using the target Language Model simultaneously. 
The target LLM evaluates the probability or likelihood of each drafted token, enabling efficient and fast verification.

By combining drafting and parallel verification, speculative decoding can significantly speed up inference, thereby accelerating the inference of generative models.

\section{Methodology}
\noindent 
In this section, we will introduce background knowledge of the auto-regressive decoding and the speculative decoding in \ref{sec:autoregressive}, the lightweight draft model in \ref{section: lightweight draft}, and the training details in \ref{sec:model_training}. 
The general model architecture is shown in Figure \ref{fig:model}. 
\begin{figure*}
    \centering
    \zihao{5-} 
    \includegraphics[width=\textwidth]{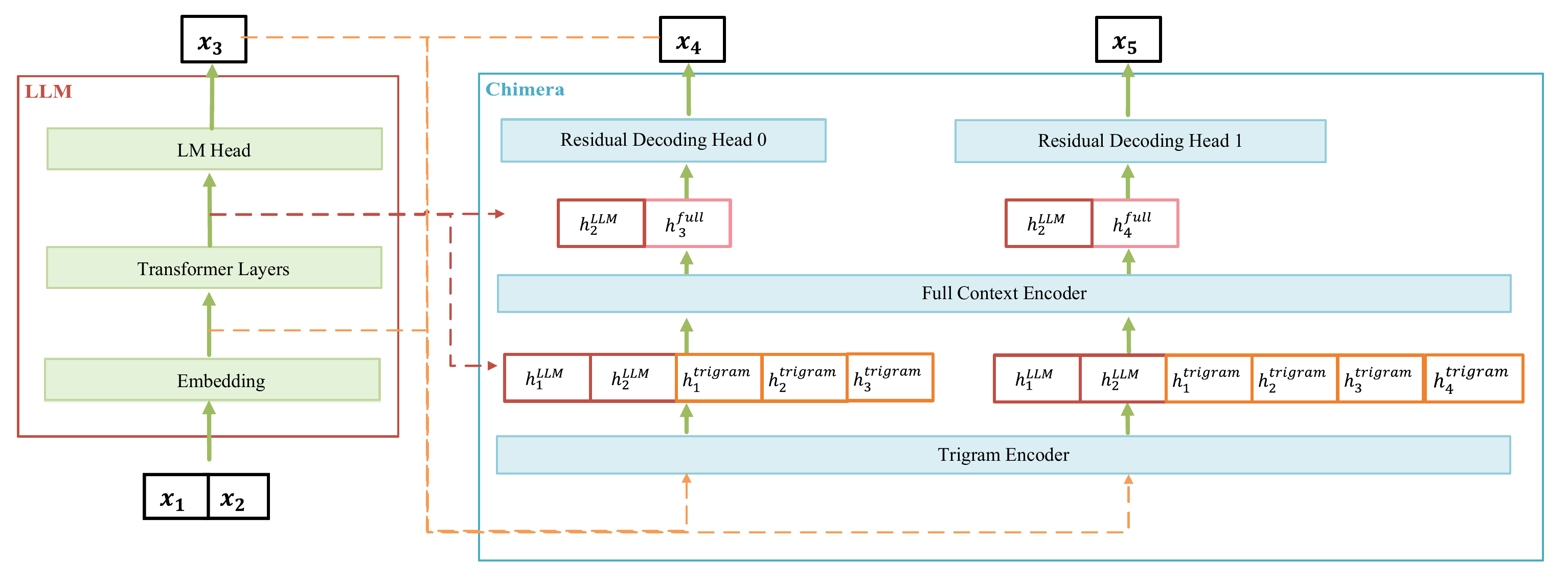}
    \caption{\textbf{The architecture of the Chimera model }. 
    The embeddings of prefix sequence are denoted as $x_1, x_2, x_3, x_4, x_5$. 
    The last hidden states are signified as $h^{LLM}_{1}$, $h^{LLM}_{2}$ .
    The trigram calculated by formula \ref{trigram_calculations} are represented by
    $h^{trigram}_{1}$, $h^{trigram}_{2}$ and the full context hidden states calculated by formula \ref{fullcontext_calculations} are signified as $h^{full}_{2}$, $h^{full}_{3}$ , with subscripts indicating their positions in the sequence .}
    \label{fig:model}
\end{figure*}

\subsection{The Lightweight Draft Model}
\label{section: lightweight draft}
\noindent 
The lightweight draft model consists of three modules: (1) trigram encoder, (2) full context encoder, and (3) residual decoding heads. 
\subsubsection{Trigram Encoder}
\noindent 
We adopt a trigram encoder as the bottom layer of the lightweight draft model to save inference time without compromising too much performance. 
As shown in previous work\upcite{jawahar2019what,rogers2020primer}, lower layers of pre-trained language models mainly focus on short-range dependencies.  
Truncating the entire input into chunks with short lengths can reduce inference time as the computational complexity is related to the input length. 
For example, the time and space complexity of self-attention is quadratic in the input length\upcite{keles2023complexity}.

The embeddings of prefix sequence are denoted as $ \mathbf{X}=\{x_1,\cdots,x_n\}$, where $n$ is the number of tokens in the input prefix sequence, $x_i \in \RM^{d}$, $d$ is the dimensionality of the embedding. 
The representation of a trigram $t_{i} \in \RM ^{3d}$ is the concatenation of embeddings and denoted as $[x_{i-2};x_{i-1};x_{i}]$. 
The input of the trigram encoder is denoted as $\mathbf{T} = \{t_1, \cdots, t_m\}$. 
Note the first two trigrams are collapsed to unigram and bigram, i.e., $t_{1} = [x_1]$ and $t_{1} = [x_1;x_2]$. 
When predicting the subsequent token in different positions, the number of trigrams in the input is different. 
We use $m$ to denote it. 
The relation between the number of tokens and the number of trigrams is described as follows: A sequence that consists of $m$ tokens can be chunked into $m$ trigrams. 

As shown in Figure \ref{fig:trigram}, the trigram encoder consists of an MLP with $2$ linear layers, parameterized by $\textbf{W}_{1} \in \RM^{3d \times d}$ and  $\textbf{W}_{2} \in \RM^{d \times d}$ respectively.
Given the input is $t_i$, the corresponding hidden states $h^{trigram}_{i}$ produced by the trigram encoder is defined as follows,
\begin{equation}
    h^{trigram}_{i} = \mathrm{MLP}(\mathrm{SiLU}(\mathrm{MLP}(t_i))),
    \label{trigram_calculations}
\end{equation}
where SiLU is the activation function. 
The output sequence of the trigram encoder is denoted as $\mathbf{H}^{trigram} = \{ h^{trigram}_{1}, \cdots, h^{trigram}_{m} \}$. 

\begin{figure}[h]
    \centering
    \zihao{5-} 
    \includegraphics[width=0.4\textwidth]{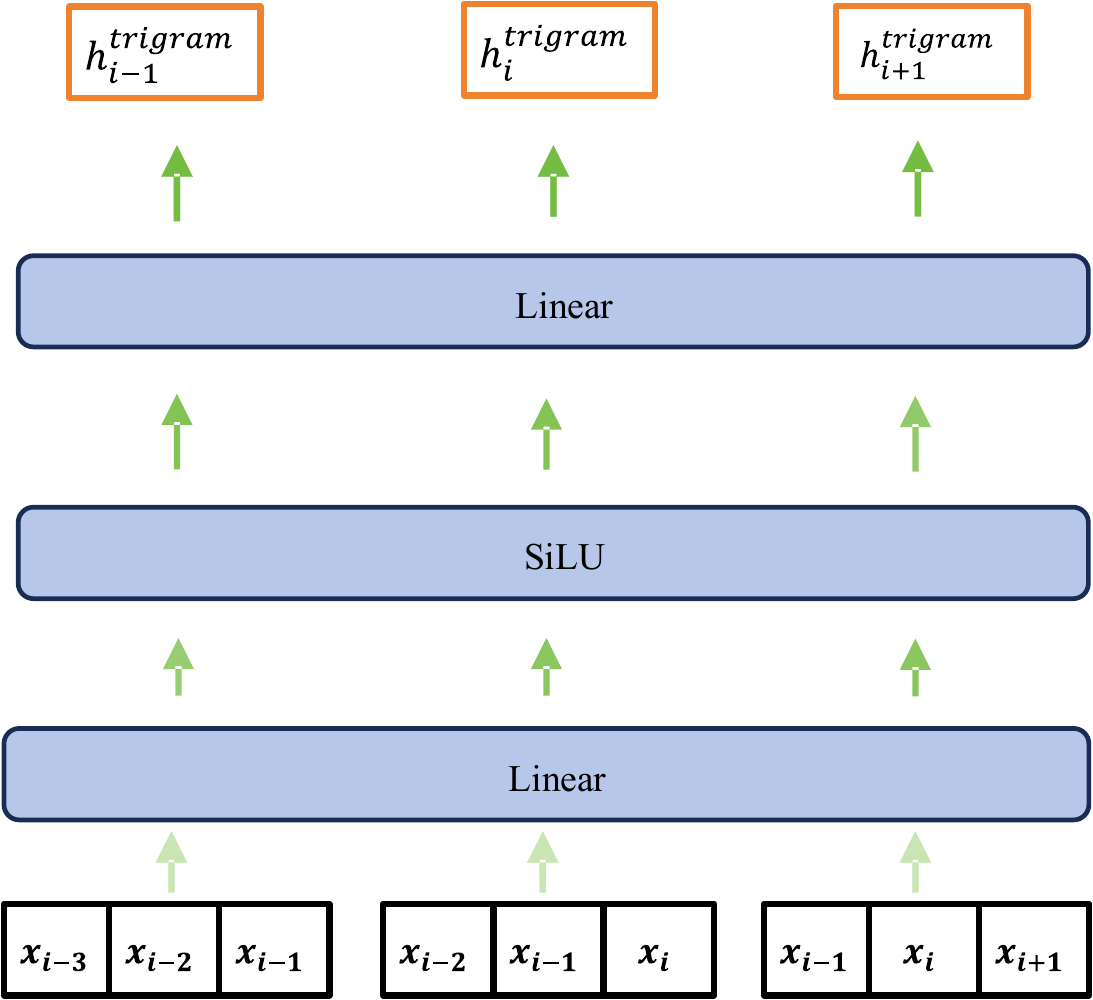}
    \caption{\textbf{The architecture of trigram encoder}. $x_{i-3}, x_{i-2} ... x_{i+1}$ represent the embeddings of token at each corresponding position. $h^{trigram}_{i-1},h^{trigram}_{i},h^{trigram}_{i+1}$ indicate the outputs of the trigram encoder.}
    \label{fig:trigram}
\end{figure}

An additional advantage of the trigram encoder is that we can retrieve the hidden states of trigrams through a pre-computed lookup table, instead of computing them on the fly, further saving inference time.
We employ a fixed-size trigram dictionary as a cache, which is loaded into memory. 
Whenever a new trigram is encountered that is not included in the dictionary, the least recently used trigram from the cache is moved to secondary storage, the corresponding entry is deleted from the cache, and the new trigram is loaded into memory. 

\subsubsection{Full Context Encoder}
\noindent 
We propose a full context encoder implemented via a Transformer block to capture the long-range dependencies. 
Capturing long-range dependencies across the entire context typically requires multiple Transformer blocks, but a lightweight draft model cannot afford that. 
Fortunately, the original LLM has properly encoded dependencies among prefix sequences. 
Therefore, we fully utilize the readily available hidden states from the original LLM.

As shown in Figure \ref{fig:Fusion attn_mask}, the input of the full context encoder, denoted as $[\mathbf{H}^{LLM}; \mathbf{H}^{trigram}]$, is the concatenation of the mature hidden states from the original LLM $\mathbf{H}^{LLM}=[h^{LLM}_{1}, \cdots, h^{LLM}_{n}]$ and the green hidden states from the trigram encoder $\mathbf{H}^{trigram}=\{h^{trigram}_{1}, \cdots, h^{trigram}_{m}\}$.  
This results in a longer input length, i.e., $m+n$.
To manage this increased length, we only retain the last $m$ hidden states as the output of the full context encoder. 
The full context encoder is implemented using a Transformer block.
The output of the full context encoder is defined as follows:
\begin{equation}
\begin{split}
        [\mathbf{H}^{discard}; \mathbf{H}^{full}]= 
        \mathrm{Transformer}([\mathbf{H}^{LLM}; \mathbf{H}^{trigram}]).
\end{split}
\label{fullcontext_calculations}
\end{equation}

During training, the attention mask is meticulously designed. 
As long as there is a future token in a trigram, then this trigram is masked. 
An example illustrating the attention mask mechanism during training is presented in Figure \ref{fig:training attnmask}.

\begin{figure}[h]
    \centering
    \includegraphics[width=\linewidth]{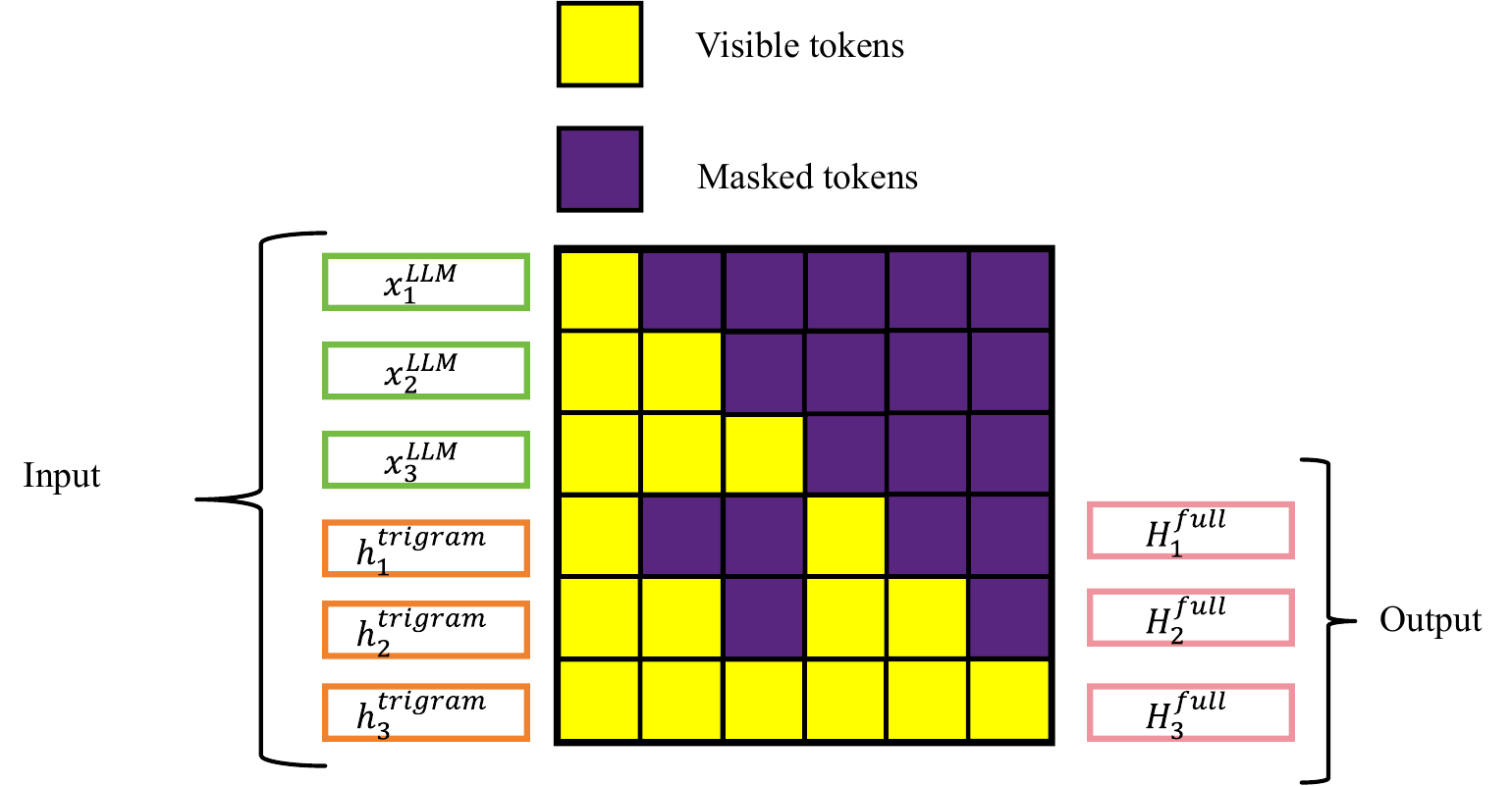}
    \caption{The provided illustration depicts the attention masks employed in the Full Context Encoder during the training phase. 
    The grey rectangles represent the tokens that should be masked during the attention calculation. 
    The input to the full context encoder is the concatenation of the last hidden states from the original language model and the outputs of the trigram encoder. }
    \label{fig:training attnmask}
\end{figure}
\begin{figure}
    \centering
    \zihao{5-} 
    \includegraphics[width=\linewidth]{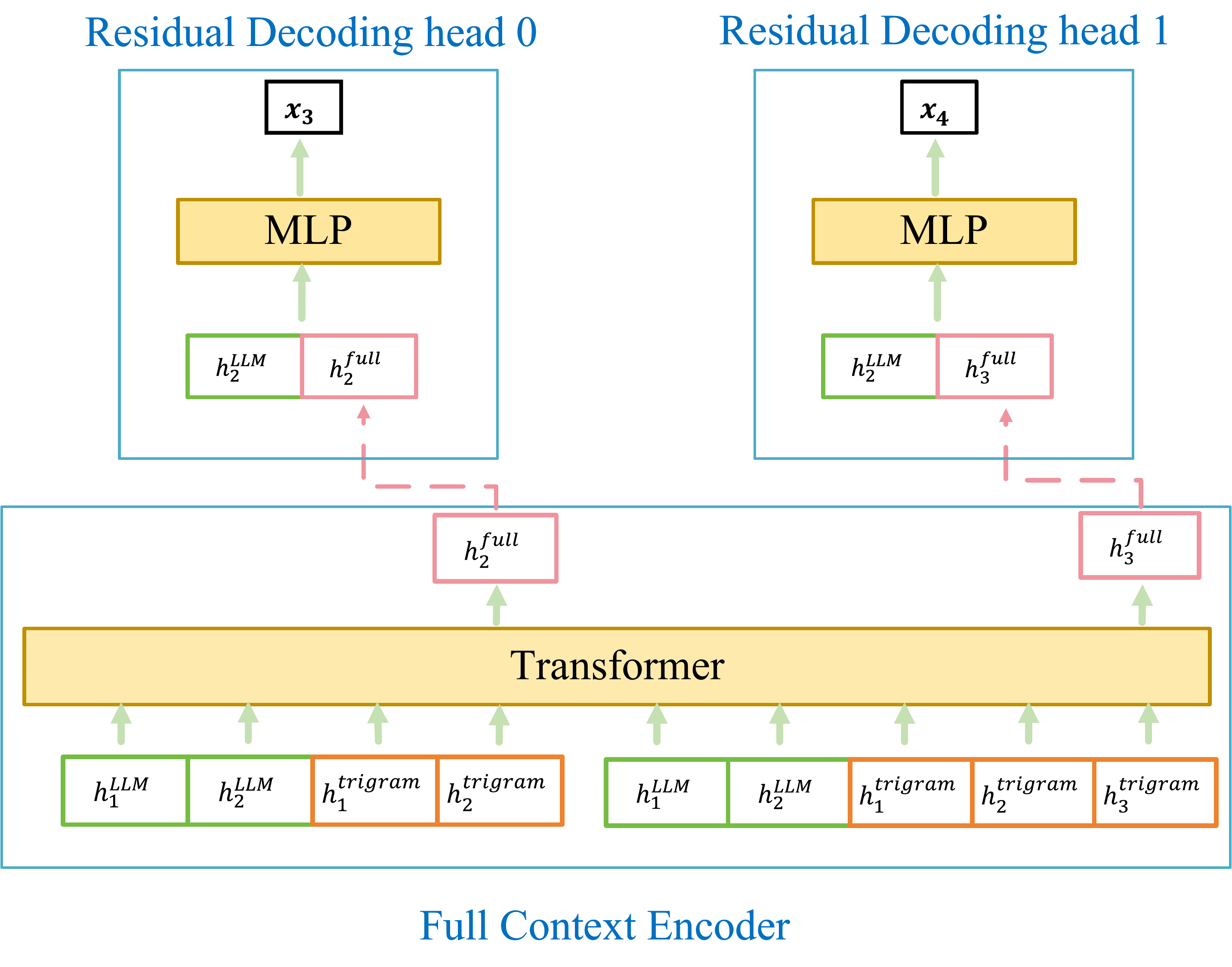}
    \caption{\textbf{The architecture of full context encoder and residual decoding heads}. 
    $h^{LLM}_{1}$ and $h^{LLM}_{2}$ represent the last hidden states of original LLM. 
    $h^{full}_{2}$ and $h^{full}_{3}$ indicate the outputs of full context encoder.}
    \label{fig:Fusion attn_mask}
\end{figure}

\subsubsection{Residual Decoding Heads}
\noindent 
We propose multiple residual decoding heads to generate subsequent words in different offset positions. 
The decoding head in the offset $k \in {0, \cdots, K-1}$ is responsible for predicting ($n+2+k$)-th word, where $K$ is the number of decoding heads in the draft model and $n$ is the number of tokens in prefix sequence. 
For example, given the prefix sequence with length $n$, the original LLM predicts the ($n+1$)-th word using its original LLM head. 
The $0$-th decoding head predicts the $(n+2)$-th word.
Due to the limited capacities of the lightweight draft model, residual decoding heads not only take the hidden state of the last token from the full context encoder as input but also consider the hidden state of the last token from the original LLM. 
The input of residual decoding head that predicts the subsequent word in the offset $k$ is the concatenation of both last hidden states $h^{LLM}_{n}$ and $h^{full}_{n+1+k}$. 

The output of the residual decoding head is defined as follows,
\begin{equation}
\begin{split}
       \mathbf{p}_{n}^{k}= 
        \mathrm{Softmax}(\mathrm{MLP}[h^{LLM}_{n}; h^{full}_{n+1+k}]), 
\end{split}
\label{residual decoding head}
\end{equation}
Where $\mathrm{MLP}$ is parameterized by $\mathrm{\Theta_{k}} \in \RM^{d \times |V|}$, and $|V|$ is the vocabulary size. 
Note that decoding heads with different offsets have different $\mathrm{MLP}$ parameters.

\subsection{Model Training}
\label{sec:model_training}
\noindent 
The training objective function consists of two parts: (1) training the draft model to predict the next word given all previous words and (2) aligning the capacity of the full context encoder to the original LLM.

To train the draft model to predict the next word correctly, we adopt the cross-entropy loss as the objective function: 
\begin{equation}
      \mathcal{L}_{\mathrm{next\_word}} =- \sum_{t=1}^{|\TM|} \sum_{k = 0}^{l} {\mathbf{y}_{t}^{k} \mathrm{log}(\mathbf{p}_{t}^{k})},
 \label{token:loss}
\end{equation}
where $l$ is the number of decoding heads in the draft model, $\mathbf{y}_{t}^{k} \in \RM^{|V|}$ is the label in a one-hot vector form where the element corresponding to the $(t+2+k)$-th word is set to one, $|\TM|$ is the total number of tokens in the training set.

To enhance the capacity of the full context encoder, we perform distillation on the hidden state of the last token of the full context encoder. 
They are trained with supervision signals from hidden states of the last layer of the original LLM. 
The distillation loss is defined as the mean-squared (MSE) error loss of hidden states between the original LLM and the full context encoder:

\begin{equation}
\mathcal{L}_{distill}=\sum_{t=1}^{|\TM|} \mathrm{MSE} (h_{t}^{full}, h_{t}^{LLM}).
\label{FE:loss}
\end{equation}

The final objective function is defined as
\begin{equation}
    \mathcal{L}  = \mathcal{L}_{next\_word} +\mathcal{L}_{distill}. 
\end{equation}

\subsection{Verification}
\noindent Building upon the Medusa approach\upcite{cai2024medusa}, we employed a tree-structured attention mechanism in our verification process. Specifically, given the tree-structured draft, the original large language model computes the probability of each candidate token through a single forward pass. 
We then leveraged both greedy decoding\upcite{kim2023speculative} and typical decoding methods to sample acceptable token sequences from the draft tokens. 
The typical decoding approach selects candidate sequences based on their probability under the original language model, rather than using rejection sampling, thus improving efficiency without sacrificing diversity.

\subsection{Inference}
\noindent The inference process is described in figure \ref{fig:inference}.During the drafting step, we can sample the top-k tokens from the three chimera heads and generate a multitude of candidate sequences leveraging these tokens. We then proceed to verify these candidate sequences in parallel with the original large language model, obtaining the conditional probability of each token within the sequences. 
For instance, we can accept the candidate sequence with the highest overall probability through a greedy search approach.
\begin{figure*}
    \centering
    \includegraphics[width = \linewidth]{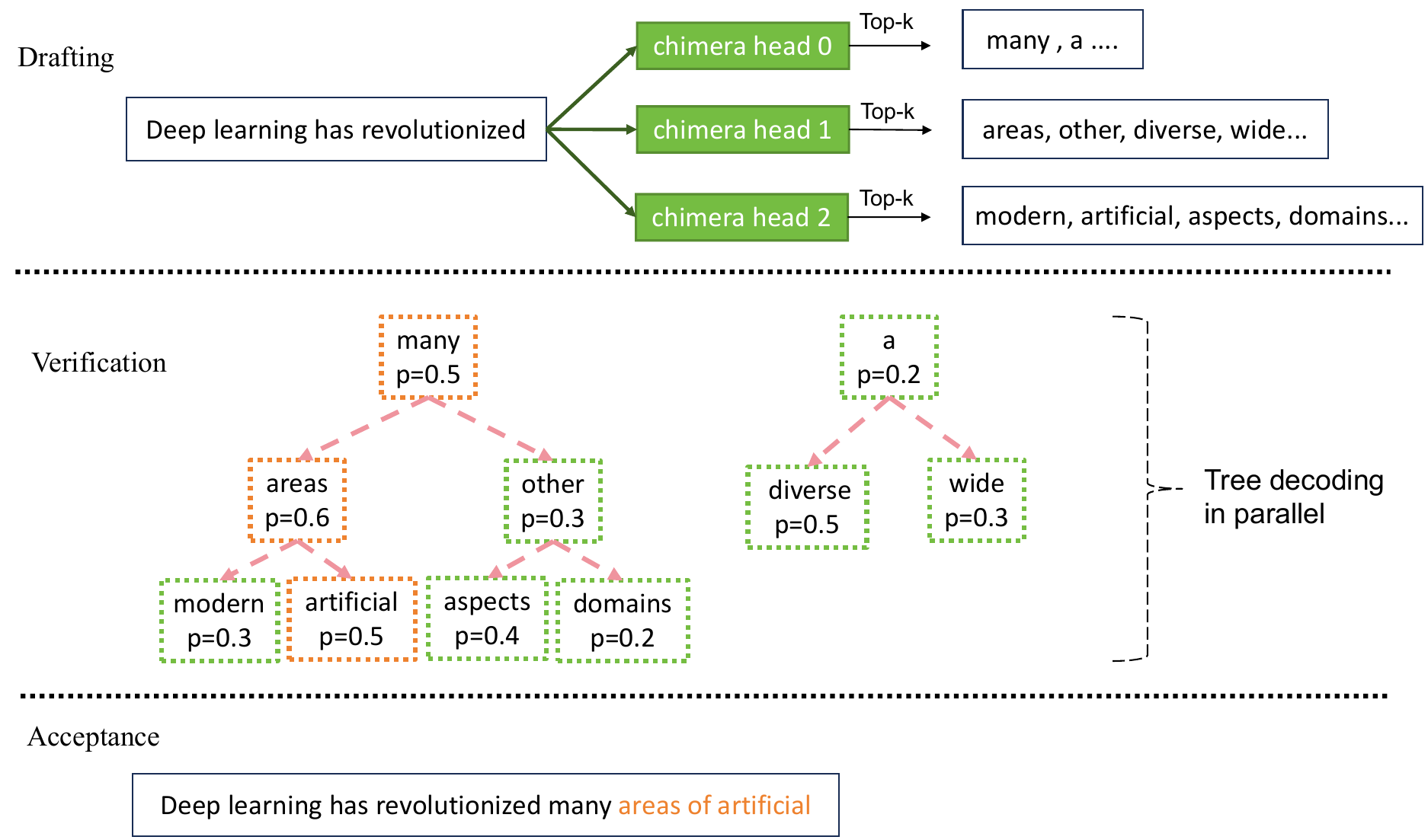}
    \caption{\textbf{Inference process}.The input sequence to the chimera model is "Deep learning has revolutionized". 
    For each position in the input sequence, the output of every chimera head is a probability distribution over the draft tokens. 
    From these distributions, we sample the top-k tokens with the highest probabilities.   }
    \label{fig:inference}
\end{figure*}

\section{Experiment}
\subsection{Experiment Setup}

\noindent All experiments are conducted on  a server with 32-core CPU, 64 GiB host memory, and a A800(80G) GPU.

\textbf{Backbone LLMs.}
We evaluate the performance of our method on various backbone LLMs, including Vicuna-7B, Vicuna-13B, Vicuna-33B\upcite{vicuna2023}, LlaMA-2-chat-7B, and LlaMA-2-chat-13B\upcite{touvron2023llama}. 
Detailed information regarding backbone LLMs can be found in Table \ref{tab:Model}.

\begin{table}[ht]
    \centering
    \caption{Summary of Models}
    \vspace{10pt} 
    \small
     \setlength{\tabcolsep}{3pt} 
    \renewcommand{\arraystretch}{1.2} 
    \begin{tabular}{lccccc}
    \toprule
       Model  &  Params &  Layer & Hidden size & Attention head\\
       \midrule
       Vicuna-7B & 7B & 32 & 4096 & 32\\
       Vicuna-13B & 13B & 40 & 5120 & 40\\
       Vicuna-33B & 33B & 60 & 6656 & 52\\
       LlaMA-2-chat-7B & 7B & 32 & 4096 & 32\\
       LlaMA-2-chat-13B & 13B & 40 & 5120 & 40\\
       \hline
    \end{tabular}
    \label{tab:Model}
\end{table}

\textbf{Datasets.} Chimera (with specific backbone LLMs) are trained on the ShareGPT dataset\upcite{aeala_sharegpt_vicuna_unfiltered}, which consists of 70,000 crafted dialogues for multi-turn conversations. 
The performances of all methods are evaluated in two benchmarks: MT-bench\upcite{zheng-2023-wku} and Vicuna-bench\upcite{vicuna2023} to evaluate multi-turn dialogue and general capabilities, respectively.

\textbf{Metrics.} Like other speculative sampling-based methods, Chimera primarily focuses on latency rather than throughput. 
We assess acceleration effects using the following metrics: 

$\bullet$ Wall-clock speedup\upcite{stern2018blockwise}: The actual  execution time acceleration ratio  relative to  autoregressive decoding. 

$\bullet$  Average acceptance length: The average number of tokens accepted per forward pass of the target LLM. 

$\bullet$ Top-k accuracy: The probability of i-th token predicted by the original LLM, denoted as $t^{LLM}_{i}$, contained within the top-k most probable tokens predicted by the Chimera model, denoted as $t^{chimera}_{i1}, t^{chimera}_{i2}, ..., t^{chimera}_{ik}$.

Formally, top-k accuracy of i-th token can be expressed as:
\begin{equation}
     Acc_{ik} = \mathbb{P}\left(t^{LLM}_{i} \in \{{t^{chimera}_{i1}, t^{chimera}_{i2}, ..., t^{chimera}_{ik}}\}\right)
\end{equation}

\textbf{Decoding strategies.}
Two decoding strategies are tested for verification: greedy decoding and typical decoding\upcite{cai2024medusa}. 
For typical decoding, the temperature is set as $T=1$.

\subsection{Baselines}

\noindent Our approach is compared against three methods, including the default autoregressive decoding, and two recent methods for LLM acceleration.

\textbf{Lookahead}\upcite{fu2024break}. 
Lookahead Decoding parallelly generates multiple disjoint $n$-grams to accommodate the future parts of a sequence. 
It reformulate the autoregressive decoding as the solution of nonlinear equation systems. 
Then, Jacobi iteration is employed for parallel decoding, capturing, validating, and integrating the generated $n$-grams into the sequence.

\textbf{Medusa-1}\upcite{cai2024medusa}.
Medusa-1
is a parameter-efficient training approach with multiple decoding heads to be trained on the same model. 
By relaxing the requirement for matching the distribution of the original model, Medusa enables faster and non-greedy generation and addresses the challenges of speculative decoding.

Medusa-2 requires fine-tuning of the backbone LLMs, which is distinct from the approach proposed in this paper. 
In the subsequent sections, we will consistently refer to Medusa-1 as "Medusa" for brevity.




\begin{figure*}[h]
    \centering
    \zihao{5-} 
    \includegraphics[width=\textwidth]{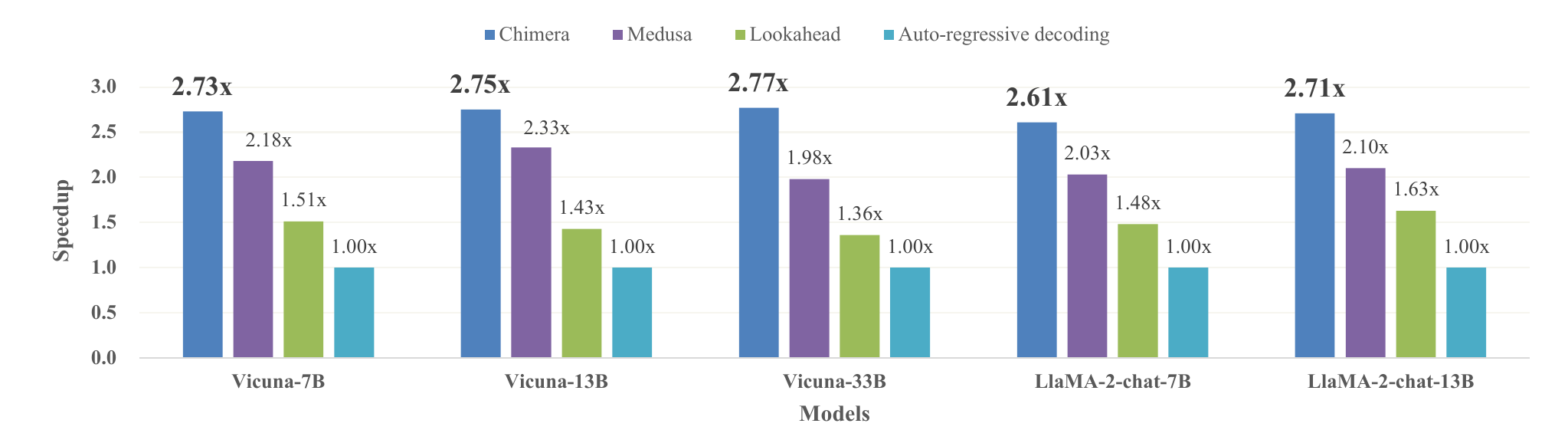}
    \caption{The wall-clock speedup of inference latency on the MT-bench for Vicuna and LlaMA-2-chat models under the greedy decoding(temperature = 0) setting. }
    \label{fig:modelspeedup}
\end{figure*}

\subsection{Main results}

\subsubsection{Accelearting LLM inference}

\noindent In order to demonstrate the efficacy of our methods, we present the speedup ratio, which is the ratio between the original inference time and the accelerated inference time, as well as the average accept length. 
The average accept length corresponds to the average number of tokens that are accepted during each forward pass of the backbone LLMs.

Figure~\ref{fig:modelspeedup} demonstrated that our method achieved better acceleration across all five backbone LLMs on MT-bench with greedy decoding. 
The maximum speedup ratio was 2.77x, surpassing Medusa by 0.79x and Lookahead by 1.41x. The average speedup ratio was 2.77x.


\textbf{Study in different decoding methods.}
Table~\ref{tab:MT-bench} also test the typical decoding setting (T=1) on MT-bench, where our method exhibited an even better acceleration on greedy decoding. 
Specifically, the best speedup ratio is observed on Vicuna-33B, attaining a maximum speedup of 2.91x. 
The minimum speedup was 2.62x, with an average speedup of 2.79x, surpassing greedy search by 0.14x in terms of speed. The maximum average length was 3.38, the minimum was 3.30, and the overall average was 3.32.

\textbf{Study in different datasets.}
Table~\ref{tab:Vicuna-bench} confirms all findings in MT-bench to Vicuna-bench, demonstrating that our approach can be applied to various domains. 
Our approach demonstrated the best acceleration performance on Vciuna33B, achieving a maximum speedup of 2.77x, under the greedy decoding method. 
The minimum speedup across all five models was 2.61x, with an average speedup of 2.72x. 
The maximum average accept length was 3.30, indicating that, on average, the model could predict 3.3 tokens in one forward pass. 
The minimum average accept length was 3.21, with an overall average of 3.26.

\begin{table}[ht]
    \centering
    \caption{Wall-clock speedup and average acceptance length on 
    MT-bench, V represents Vicuna, and L2 stands for LlaMA-2-chat. }
    \vspace{10pt} 
    \resizebox{\linewidth}{!}{
    \begin{tabular}{lccc}
        \toprule 
        Decoding strategies & Model&  Speedup & Average accept length   \\ 
        \midrule 
         \multirow{6}{*}{Greedy}&V-7B& 2.73x &3.30  \\ 
          \multirow{6}{*}{ } & V-13B& 2.75x &3.26  \\ 
         \multirow{6}{*}{}& V-33B& 2.77x &3.23   \\ 
         \multirow{6}{*}{ } &L2-7B& 2.61x &3.28\\ 
          \multirow{6}{*}{ } &L2-13B& 2.74x&3.21  \\ \cline{1-4}
          \multirow{6}{*}{Typical}&V-7B& 2.81x&3.38\\ 
          \multirow{6}{*}{ } & V-13B& 2.85x &3.35 \\ 
         \multirow{6}{*}{}& V-33B& 2.91x &3.31\\
         \multirow{6}{*}{ } &L2-7B& 2.62x &3.31\\
          \multirow{6}{*}{ } &L2-13B&2.74x&3.30 \\
         \hline
    \end{tabular}
    }
    
    \label{tab:MT-bench}
\end{table}

\begin{table}[ht]
    \centering
    \caption{Wall-clock speedup and average acceptance length on Vicuna-Bench, V represents Vicuna, and L2 stands for LlaMA-2-chat. }
    \vspace{10pt} 
    \resizebox{\linewidth}{!}{
    \begin{tabular}{lccc}
        \toprule 
        Decoding strategies& Model&  Speedup & Average accept length  \\ 
        \midrule 
         \multirow{6}{*}{Greedy}&V-7B& 2.83x &3.47 \\ 
          \multirow{6}{*}{ } & V-13B& 2.87x &3.42  \\
         \multirow{6}{*}{}& V-33B& 2.89x &3.54   \\ 
         \multirow{6}{*}{ } &L2-7B& 2.75x &3.44 \\ 
          \multirow{6}{*}{ } &L2-13B& 2.77x&3.41 \\ \cline{1-4}
          \multirow{6}{*}{Typical}&V-7B& 2.85x &3.49\\ 
          \multirow{6}{*}{ } & V-13B& 2.88x &3.45 \\
         \multirow{6}{*}{}& V-33B& 2.90x &3.45\\
         \multirow{6}{*}{ } &L2-7B& 2.85x &3.37 \\ 
          \multirow{6}{*}{ } &L2-13B&2.80x &3.34\\ 

         \hline
    \end{tabular}
    }
    
    \label{tab:Vicuna-bench}
\end{table}

\textbf{Case study in different domains.}
We evaluated the performance of our method using the greedy decoding technique on 13 different cases based on the Vicuna-7B model. 
The results are presented in Figure \ref{fig:casespeedup} and Table \ref{tab:different kinds of situation}. 
Our model performed well in the coding and roleplay cases, achieving speedup ratios of 3.03x and 2.91x, respectively. These speedup ratios surpassed the average speedup of Vicuna-7B. 
The lowest speedup ratio observed was 2.51x, which still outperformed other methods, and the deviation from the average speedup was only 12\%. 
This indicates that our method exhibits strong generalization capabilities and does not suffer from significant performance degradation across different cases. 
\begin{figure}[ht]
    \centering
    \zihao{5-} 
    \includegraphics[width=\linewidth]{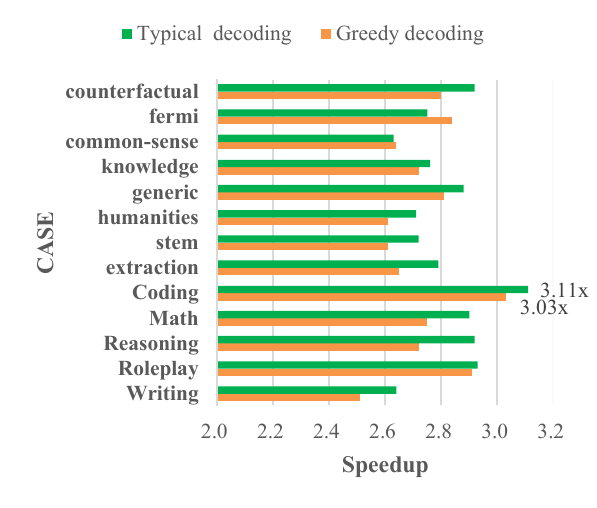}
    \caption{Wall clock speedup of different cases. These data are sourced from the Vicuna-Bench and MT-bench.} 
    \label{fig:casespeedup}
\end{figure}

\begin{table}[ht]
    \centering
    \caption{Case Test in Vicuna-7B. The metric is the average accept length.}
    \vspace{10pt} 
    \resizebox{0.7\linewidth}{!}{
    \begin{tabular}{l|cc}
        \toprule 
        \multirow{2}{*}{Case} & \multicolumn{2}{c}{Decoding strategies} \\
\cline{2-3} 
&  Greedy &  Typical\\
   
        \midrule 
        Writing & 3.31 & 3.45 \\ 
        Roleplay & 3.51 & 3.57 \\ 
        Reasoning & 3.55 & 3.61 \\ 
        Math & 3.41 & 3.66 \\ 
        Coding & 3.57 & 3.82 \\ 
        Extraction & 3.27 & 3.57 \\ 
        Stem & 3.29 & 3.45 \\ 
        Humanities & 3.28 & 3.39 \\ 
        Generic & 3.46 & 3.71 \\ 
        Knowledge & 3.43 & 3.51 \\ 
        Common-sense & 3.27 & 3.49 \\ 
        Fermi & 3.51 & 3.62 \\ 
        Counterfactual & 3.57 & 3.76 \\
        \hline 
    \end{tabular}
    }
    \label{tab:different kinds of situation}
\end{table}

\subsubsection{Accurate decoding heads}

\noindent The accuracy of each decoding head is in Table \ref{tab:head_acc}. 
Our method outperformed Medusa with an average margin of 0.13 on Medusa head 0 and achieved a maximum improvement of 27\% on top1 compared to Medusa head 0. 
On Medusa head 1, Medusa head 2, and Medusa head 3, Chimera surpassed Medusa by 79\%, 217\%, and 267\% on the top1 metric, respectively. 
These results demonstrate that our model exhibits higher accuracy in predicting longer sequences compared to Medusa. This also suggests that our model is better able to extract the necessary information for predicting longer sequences effectively.
\begin{table}[ht]
    \centering
    \caption{Head Accuracies of Medusa and Chimera in Vicuna-7B. Top-k represents the top-k accuracy of model.}
    \vspace{10pt} 
    \resizebox{0.45\textwidth}{!}{\begin{tabular}{lcccccc}
        \toprule 
         
        Head  & Model& Top-1  &  Top-2 & Top-3 & Top-4 & Top-5\\ 
        \midrule 
         \multirow{2}{*}{0}&Chimera& 0.66&0.78&0.83 &0.87 &\textbf{0.90}\\ 
          \multirow{2}{*}{ } & Medusa&0.52 & 0.65 & 0.71 & 0.74 & 0.77 \\ \cline{2-7}\cline{1-1}
       \multirow{2}{*}{1}& Chimera&0.49&0.55&0.60 &0.68 &\textbf{0.75}\\
         \multirow{2}{*}{ } &Medusa&0.29 & 0.39 &0.45 & 0.50 &0.53 \\ \cline{2-7}\cline{1-1}
         \multirow{2}{*}{2}&  Chimera& 0.38&0.44 &0.49 &0.55 &\textbf{0.62}\\
         \multirow{2}{*}{ } &Medusa&0.12 & 0.25 &0.31 & 0.34 &0.38 \\\cline{2-7}\cline{1-1}
         \multirow{2}{*}{3}&  Chimera&0.33&0.38&0.43 &0.46 &\textbf{0.54}\\
         \multirow{2}{*}{ } &Medusa&0.09 & 0.18 &0.23 & 0.27 &0.30 \\
         
         \hline
    \end{tabular}
    }
   
    \label{tab:head_acc}
\end{table}

\subsection{Ablation Study}

\subsubsection{Trigram encoder}

          

\noindent\textbf{Trigram cache.}
To reduce computation time, a preconstructed trigram cache, which stores the outputs of commonly encountered N-grams after passing through the trigram encoder, can be utilized. On Vicuna-7B, this approach can result in a maximum speedup of 1.45x compared to not using the trigram cache. Specifically, the model's acceleration improves from 2.69x to 2.81x under the greedy decoding method.


\textbf{Trigram encoder.}
The selection and ablation study of the trigram encoder is summarized in the table \ref{tab:N-gramablation}: 
In terms of accuracy, the performance of the different heads in 1-gram, 2-gram, 3-gram, 4-gram, and 5-gram consistently surpasses that of the 1-gram and 2-gram heads by less than 1\%. 
Notably, the 3-gram head exhibits relatively faster computation and is better suited for the n-gram cache strategy. 
Consequently, Chimera ultimately selects the 3-gram approach. 
Additionally, it is observed that using the frozen transformer yields inferior results compared to using the n-gram approach. 
On the other hand, the performance of the transformer (finetune) is comparable to that of the 3-gram, but the transformer requires significantly more computation time and is not compatible with the cache strategy. 
In practice, the acceleration effect is still inferior to that of the N-gram Encoder.
\begin{table}[h]
    \centering
    \caption{Impact of $k$-gram encoder on Chimera. Chimera ($k$-gram) indicates the Chimera model with a $k$-gram encoder. 
    Transformer (Frozen) represents that we use the frozen first transformer layer of the original LLM, and Transformer (Finetune)  is one layer of transformer finetuned, the metric is the prediction accuracy of top-5 tokens.}
    \vspace{10pt} 
    \resizebox{\linewidth}{!}{\begin{tabular}{l|cccc}
        \toprule 
        \multirow{2}{*}{Model} & \multicolumn{4}{c}{Head} \\
\cline{2-5} 
&  0&  1&2&3\\

        \midrule 
        Chimera (1-gram) &0.83&0.65&0.50 &0.48 \\ 
        Chimera (2-gram) &0.87&0.71&0.58 &0.51 \\
        Chimera (3-gram) &0.90&0.75&0.62 &0.54\\ 
        Chimera (4-gram)&0.91 & 0.75 & 0.63 & 0.54  \\ 
        Chimera (5-gram) &0.91 &  0.76 & 0.63 & 0.54 \\ \cline{1-5}
        Transformer (Frozen) &0.85 & 0.60  & 0.49 & 0.41 \\ 
        Transformer (Finetune) &0.89 & 0.74 & 0.56 & 0.52 \\ 
         \hline
    \end{tabular}
    }
    \label{tab:N-gramablation}
\end{table}



\subsubsection{Residual connection of decoding heads}
\noindent The effect of removing the residual connection is shown in Figure \ref{fig:residualablation}. 
It can be observed that the performance without residual is similar to Medusa, but the inclusion of residual connections significantly improves the accuracy of each head. 
Specifically, compared to the model without residual connections, head0 shows an improvement of 0.12, while head3 outperforms Medusa by 0.26. 
This suggests that the residual head is more effective in predicting tokens for longer sequences in the LLM context.
\begin{figure}[h]
    \centering
    \zihao{5-} 
    \includegraphics[width=0.9\linewidth]{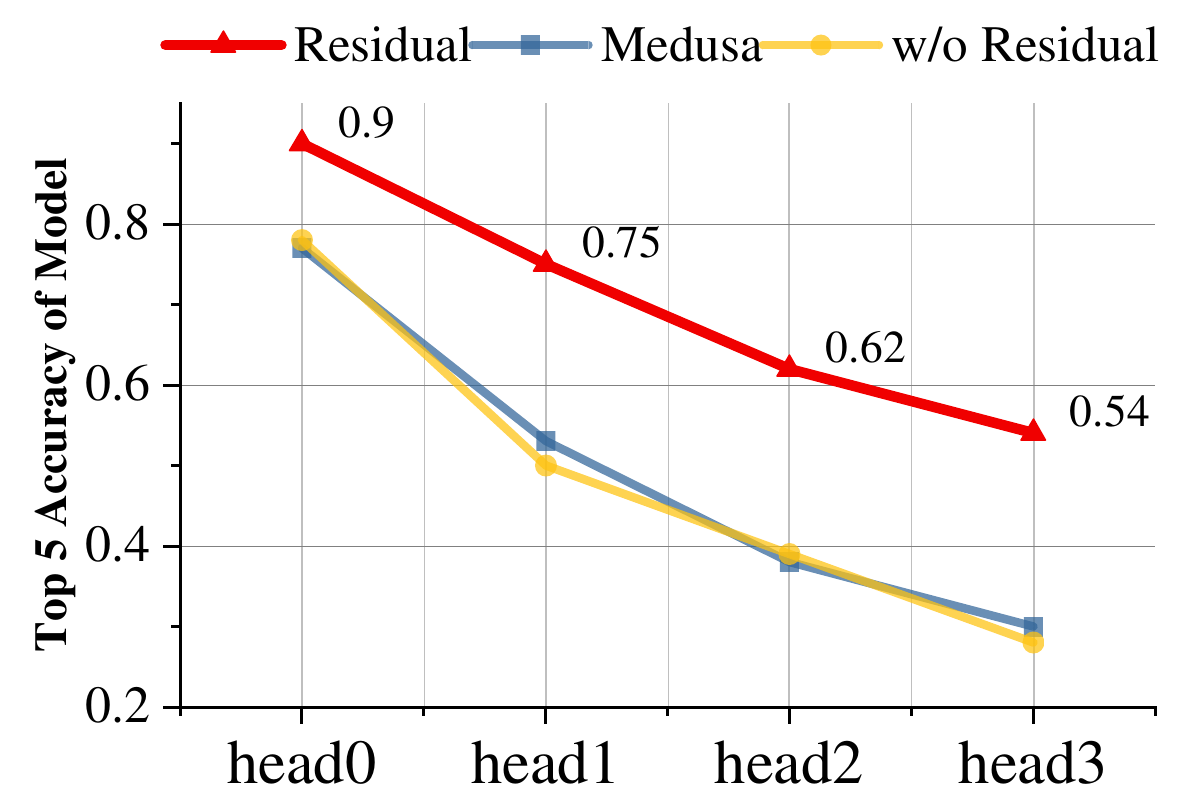}
    \caption{The prediction accuracy of top-5 tokens in different variants of residual decoding heads. 
    The residual approach represents a chimera model that utilizes residual decoding heads, whereas the "without residual" (w/o residual) variant implies that the input to the heads is solely the output of the full-context encoder, without any residual connections. }
    \label{fig:residualablation}
\end{figure}
       

\section{Conclusion}
\noindent This paper presents Chimera, a novel framework for speculative sampling. 
We propose a lightweight draft model that leverages previously generated tokens to predict subsequent words. 
To ensure accuracy and efficiency, we introduce two strategies in the lightweight draft model: capturing short-range dependencies at the bottom layer and utilizing readily available representations from the original LLM.
In experiments conducted on the Vicuna and LlaMA-2-chat series, Chimera achieves an average latency speedup ratio of 2.7x compared to auto-regressive decoding. 


\vspace{2mm}
\noindent
\textbf{\zihao{5}Acknowledgement}
\vspace{2mm}

\noindent This research was supported by the Guangzhou Basic and Applied Basic Research Foundation (Grant No. 2023A04J1687), National Natural Science Foundation of China (Grant No. 6230070401), South China University of Technology-TCL Technology Innovation Fund, the Postdoctoral Fellowship Program of CPSF (Grant No. GZC20230841). 

\vskip 2mm
\renewcommand\refname{\zihao{5}

\vskip .2mm
{\zihao{5-}
\textbf{
\hspace{-5mm}
\scalebox{1}[1.0]{\makebox[5.6cm][s]{%
I\hspace{0.70pt}S\hspace{0.70pt}S\hspace{0.70pt}N\hspace{0.70pt}{\color{white}%
l\hspace{0.70pt}l\hspace{0.70pt}}1\hspace{0.70pt}0\hspace{0.70pt}0\hspace{0.70pt%
}7\hspace{0.70pt}-\hspace{0.70pt}0\hspace{0.70pt}2\hspace{0.70pt}1\hspace{0.70pt%
}4\hspace{0.70pt}{\color{white}l\hspace{0.70pt}l\hspace{0.70pt}}0\hspace{0.70pt}%
?\hspace{0.70pt}/\hspace{0.70pt}?\hspace{0.70pt}?\hspace{0.70pt}{\color{white}%
l\hspace{0.70pt}l\hspace{0.70pt}}p\hspace{0.70pt}p\hspace{0.70pt}?\hspace{0.70pt}?\hspace{0.70pt}?%
-\hspace{ 0.70pt}?\hspace{0.70pt}?\hspace{0.70pt}?}}}

\vskip .2mm\noindent
{\zihao{5-}\textbf{\scalebox{1}[1.0]{\makebox[5.6cm][s]{%
V\hspace{0.2pt}o\hspace{0.2pt}l\hspace{0.2pt}u\hspace{0.2pt}m\hspace{0.2pt}%
e\hspace{0.4em}2\hspace{0.2pt}2,\hspace{0.4em}N\hspace{0.2pt}u\hspace{0.2pt}%
m\hspace{0.2pt}b\hspace{0.2pt}e\hspace{0.2pt}r\hspace{0.4em}1,\hspace{0.4em}%
F\hspace{0.2pt}e\hspace{0.2pt}b\hspace{0.2pt}r\hspace{0.2pt}u\hspace{0.2pt}%
a\hspace{0.2pt}r\hspace{0.2pt}y\hspace{0.4em}2\hspace{0.2pt}0\hspace{0.2pt}1\hspace{0.2pt}7}}}}


\textbf{References}}
\vskip 2mm

\begin{thebibliography}{99}
\zihao{5-} \addtolength{\itemsep}{-1em}
\vspace {1.5mm}

\bibitem{almazrouei2023falcon}
E.~Almazrouei, H.~Alobeidli, A.~Alshamsi, A.~Cappelli, R.~Cojocaru, M.~Debbah, {\'E}.~Goffinet, D.~Hesslow, J.~Launay, Q.~Malartic \emph{et~al.}, ``The falcon series of open language models,'' \emph{arXiv preprint arXiv:2311.16867}, 2023.

\bibitem{touvron2023llama2}
H.~Touvron, L.~Martin, K.~Stone, P.~Albert, A.~Almahairi, Y.~Babaei, N.~Bashlykov, S.~Batra, P.~Bhargava, S.~Bhosale \emph{et~al.}, ``Llama 2: Open foundation and fine-tuned chat models,'' \emph{arXiv preprint arXiv:2307.09288}, 2023.

\bibitem{achiam2023gpt4}
J.~Achiam, S.~Adler, S.~Agarwal, L.~Ahmad, I.~Akkaya, F.~L. Aleman, D.~Almeida, J.~Altenschmidt, S.~Altman, S.~Anadkat \emph{et~al.}, ``Gpt-4 technical report,'' \emph{arXiv preprint arXiv:2303.08774}, 2023.

\bibitem{anil2023palm}
R.~Anil, A.~M. Dai, O.~Firat, M.~Johnson, D.~Lepikhin, A.~Passos, S.~Shakeri, E.~Taropa, P.~Bailey, Z.~Chen \emph{et~al.}, ``Palm 2 technical report,'' \emph{arXiv preprint arXiv:2305.10403}, 2023.

\bibitem{bai2023qwen}
J.~Bai, S.~Bai, Y.~Chu, Z.~Cui, K.~Dang, X.~Deng, Y.~Fan, W.~Ge, Y.~Han, F.~Huang \emph{et~al.}, ``Qwen technical report,'' \emph{arXiv preprint arXiv:2309.16609}, 2023.

\bibitem{yang2023baichuan}
A.~Yang, B.~Xiao, B.~Wang, B.~Zhang, C.~Bian, C.~Yin, C.~Lv, D.~Pan, D.~Wang, D.~Yan \emph{et~al.}, ``Baichuan 2: Open large-scale language models,'' \emph{arXiv preprint arXiv:2309.10305}, 2023.

\bibitem{Leviathan2023fast}
Y.~Leviathan, M.~Kalman, and Y.~Matias, ``Fast inference from transformers via speculative decoding,'' in \emph{ICML}, vol. 202, 2023, pp. 19\,274--19\,286.

\bibitem{stern2018blockwise}
M.~Stern, N.~Shazeer, and J.~Uszkoreit, ``Blockwise parallel decoding for deep autoregressive models,'' in \emph{NeurIPS}, 2018, pp. 10\,107--10\,116.

\bibitem{chen2023accelerating}
C.~Chen, S.~Borgeaud, G.~Irving, J.-B. Lespiau, L.~Sifre, and J.~Jumper, ``Accelerating large language model decoding with speculative sampling,'' 2023.

\bibitem{cai2024medusa}
T.~Cai, Y.~Li, Z.~Geng, H.~Peng, J.~D. Lee, D.~Chen, and T.~Dao, ``Medusa: Simple llm inference acceleration framework with multiple decoding heads,'' \emph{arXiv preprint arXiv: 2401.10774}, 2024.

\bibitem{xia2024unlocking}
H.~Xia, Z.~Yang, Q.~Dong, P.~Wang, Y.~Li, T.~Ge, T.~Liu, W.~Li, and Z.~Sui, ``Unlocking efficiency in large language model inference: A comprehensive survey of speculative decoding,'' \emph{arXiv preprint arXiv:2401.07851}, 2024.

\bibitem{jawahar2019what}
G.~Jawahar, B.~Sagot, and D.~Seddah, ``What does {BERT} learn about the structure of language?'' in \emph{ACL}, 2019, pp. 3651--3657.

\bibitem{rogers2020primer}
A.~Rogers, O.~Kovaleva, and A.~Rumshisky, ``A primer in bertology: What we know about how {BERT} works,'' \emph{TACL}, vol.~8, pp. 842--866, 2020.

\bibitem{xia2023speculative}
H.~Xia, T.~Ge, P.~Wang, S.~Chen, F.~Wei, and Z.~Sui, ``Speculative decoding: Exploiting speculative execution for accelerating seq2seq generation,'' in \emph{EMNLP}, 2023, pp. 3909--3925.

\bibitem{miao2024specinfer}
X.~Miao, G.~Oliaro, Z.~Zhang, X.~Cheng, Z.~Wang, Z.~Zhang, R.~Y.~Y. Wong, A.~Zhu, L.~Yang, X.~Shi, C.~Shi, Z.~Chen, D.~Arfeen, R.~Abhyankar, and Z.~Jia, ``Specinfer: Accelerating generative large language model serving with tree-based speculative inference and verification,'' 2024.

\bibitem{kim2023speculative}
S.~Kim, K.~Mangalam, S.~Moon, J.~Malik, M.~W. Mahoney, A.~Gholami, and K.~Keutzer, ``Speculative decoding with big little decoder,'' in \emph{NeurIPS}, 2023.

\bibitem{zhou2023distillspec}
Y.~Zhou, K.~Lyu, A.~S. Rawat, A.~K. Menon, A.~Rostamizadeh, S.~Kumar, J.-F. Kagy, and R.~Agarwal, ``Distillspec: Improving speculative decoding via knowledge distillation,'' 2023.

\bibitem{spector2023accelerating}
B.~Spector and C.~Re, ``Accelerating llm inference with staged speculative decoding,'' 2023.

\bibitem{sun2024spectr}
Z.~Sun, A.~T. Suresh, J.~H. Ro, A.~Beirami, H.~Jain, and F.~X. Yu, ``Spectr: Fast speculative decoding via optimal transport,'' in \emph{NeurIPS}, 2023.

\bibitem{chen2023cascade}
Z.~Chen, X.~Yang, J.~Lin, C.~Sun, J.~Huang, and K.~C.-C. Chang, ``Cascade speculative drafting for even faster llm inference,'' 2023.

\bibitem{santilli-etal-2023-accelerating}
A.~Santilli, S.~Severino, E.~Postolache, V.~Maiorca, M.~Mancusi, R.~Marin, and E.~Rodol{\`{a}}, ``Accelerating transformer inference for translation via parallel decoding,'' in \emph{ACL}, 2023, pp. 12\,336--12\,355.

\bibitem{hooper2024speed}
C.~Hooper, S.~Kim, H.~Mohammadzadeh, H.~Genc, K.~Keutzer, A.~Gholami, and S.~Shao, ``Speed: Speculative pipelined execution for efficient decoding,'' 2024.

\bibitem{yang2023predictive}
S.~Yang, G.~Lee, J.~Cho, D.~Papailiopoulos, and K.~Lee, ``Predictive pipelined decoding: A compute-latency trade-off for exact llm decoding,'' 2023.

\bibitem{zhang2023draft}
J.~Zhang, J.~Wang, H.~Li, L.~Shou, K.~Chen, G.~Chen, and S.~Mehrotra, ``Draft $\&$ verify: Lossless large language model acceleration via self-speculative decoding,'' 2023.

\bibitem{monea2023pass}
G.~Monea, A.~Joulin, and E.~Grave, ``Pass: Parallel speculative sampling,'' 2023.

\bibitem{keles2023complexity}
F.~D. Keles, P.~M. Wijewardena, and C.~Hegde, ``On the computational complexity of self-attention,'' in \emph{ALT}, vol. 201, 2023, pp. 597--619.

\bibitem{vicuna2023}
\BIBentryALTinterwordspacing
W.-L. Chiang, Z.~Li, Z.~Lin, Y.~Sheng, Z.~Wu, H.~Zhang, L.~Zheng, S.~Zhuang, Y.~Zhuang, J.~E. Gonzalez, I.~Stoica, and E.~P. Xing, ``Vicuna: An open-source chatbot impressing gpt-4 with 90\%* chatgpt quality,'' March 2023. [Online]. Available: \url{https://lmsys.org/blog/2023-03-30-vicuna/}
\BIBentrySTDinterwordspacing

\bibitem{touvron2023llama}
H.~Touvron, L.~Martin, K.~Stone, P.~Albert, A.~Almahairi, Y.~Babaei, N.~Bashlykov, S.~Batra, P.~Bhargava, S.~Bhosale, D.~Bikel, L.~Blecher, C.~C. Ferrer, M.~Chen, G.~Cucurull, D.~Esiobu, J.~Fernandes, J.~Fu, W.~Fu, B.~Fuller, C.~Gao, V.~Goswami, N.~Goyal, A.~Hartshorn, S.~Hosseini, R.~Hou, H.~Inan, M.~Kardas, V.~Kerkez, M.~Khabsa, I.~Kloumann, A.~Korenev, P.~S. Koura, M.-A. Lachaux, T.~Lavril, J.~Lee, D.~Liskovich, Y.~Lu, Y.~Mao, X.~Martinet, T.~Mihaylov, P.~Mishra, I.~Molybog, Y.~Nie, A.~Poulton, J.~Reizenstein, R.~Rungta, K.~Saladi, A.~Schelten, R.~Silva, E.~M. Smith, R.~Subramanian, X.~E. Tan, B.~Tang, R.~Taylor, A.~Williams, J.~X. Kuan, P.~Xu, Z.~Yan, I.~Zarov, Y.~Zhang, A.~Fan, M.~Kambadur, S.~Narang, A.~Rodriguez, R.~Stojnic, S.~Edunov, and T.~Scialom, ``Llama 2: Open foundation and fine-tuned chat models,'' 2023.

\bibitem{aeala_sharegpt_vicuna_unfiltered}
Aeala, ``Sharegpt\_vicuna\_unfiltered,'' \url{https://huggingface.co/datasets/Aeala/ShareGPT_Vicuna_unfiltered}, 2022.

\bibitem{zheng-2023-wku}
Q.~Zheng, ``Wku{\_}nlp at semeval-2023 task 9: Translation augmented multilingual tweet intimacy analysis,'' in \emph{ACL}, 2023, pp. 1525--1530.

\bibitem{fu2024break}
Y.~Fu, P.~Bailis, I.~Stoica, and H.~Zhang, ``Break the sequential dependency of llm inference using lookahead decoding,'' 2024.

\bibitem{liu2023online}
X.~Liu, L.~Hu, P.~Bailis, I.~Stoica, Z.~Deng, A.~Cheung, and H.~Zhang, ``Online speculative decoding,'' 2023.

\end{thebibliography}
\end{document}